\documentclass{article}
\usepackage{float}
 \usepackage{adjustbox} 
\usepackage{graphicx} 
\usepackage{setspace}
\usepackage{todonotes}
\usepackage{authblk}
\usepackage{hyperref}
\usetikzlibrary{shapes.geometric, arrows}
\usepackage{tikz}
\usetikzlibrary{positioning}
\usepackage{amsmath}
\usepackage{caption}
\usepackage{comment}
\usepackage{color}
\usepackage{xcolor}
\usepackage{tabularx}
\definecolor{asparagus}{rgb}{0.53, 0.66, 0.42}

\title{Decoding Rarity: Large Language Models in the Diagnosis of Rare Diseases}
\author{Valentina Carbonari \footnote{University of Catanzaro}, Pierangelo Veltri \footnote{University of Calabria}, Pietro Hiram Guzzi \footnote{University of Catanzaro}}

\date{December 2024}

\begin{document}

\maketitle
\begin{abstract}

Recent advances in artificial intelligence, particularly large language models (LLMs), have shown promising capabilities in transforming rare disease research. This survey paper explores the integration of LLMs in the analysis of rare diseases, highlighting significant strides and pivotal studies that leverage textual data to uncover insights and patterns critical for diagnosis, treatment, and patient care. While current research predominantly employs textual data, the potential for multimodal data integration—combining genetic, imaging, and electronic health records—stands as a promising frontier. We review foundational papers that demonstrate the application of LLMs in identifying and extracting relevant medical information, simulating intelligent conversational agents for patient interaction, and enabling the formulation of accurate and timely diagnoses. Furthermore, this paper discusses the challenges and ethical considerations inherent in deploying LLMs, including data privacy, model transparency, and the need for robust, inclusive data sets. As part of this exploration, we present a section on experimentation that utilizes multiple LLMs alongside structured questionnaires, specifically designed for diagnostic purposes in the context of different diseases. We conclude with future perspectives on the evolution of LLMs towards truly multimodal platforms, which would integrate diverse data types to provide a more comprehensive understanding of rare diseases, ultimately fostering better outcomes in clinical settings
\end{abstract}

\section{Introduction}
Rare diseases are officially defined by the European Union as conditions affecting fewer than one in 2,000 individuals\cite{eu_comm_rare_disease_definition}, and by the United States as those impacting fewer than 200,000 people nationwide at any given time\cite{us_fda_rare_disease_definition}. Yet, collectively, rare diseases are anything but rare: over 7,000 distinct disorders have been identified, affecting an estimated 3.5–5.9

Despite this scale, patients often endure prolonged, complex diagnostic journeys. The low prevalence and clinical heterogeneity of these conditions frequently result in misdiagnoses or delayed recognition, as symptoms can mimic more common diseases or manifest atypically\cite{schieppati}. While diagnostic technologies and awareness have advanced, therapeutic development lags far behind. Today, only a small fraction of rare diseases have approved, effective treatments, leaving most patients without viable medical options\cite{faye2024time}.

This landscape presents profound challenges for clinicians, researchers, and healthcare systems alike. From a computational perspective, key hurdles include the inherent scarcity of patient data and the fragmentation of knowledge across diverse and siloed sources. Nevertheless, computer science has become a crucial ally. Bioinformatics and machine learning methods have facilitated the discovery of novel biomarkers, enhancing diagnostic precision and enabling better patient stratification\cite{rao2017rare, nguyen2022deep}. Meanwhile, in silico drug discovery and repurposing efforts have accelerated therapeutic identification, a critical advance given the limited financial incentives for orphan drug development\cite{corsello2017drug, yang2019machine}.

More recently, artificial intelligence, particularly the advent of large language models (LLMs), has opened new horizons for rare disease research\cite{shool2025systematic}. These models, trained on extensive biomedical corpora, demonstrate remarkable ability to understand, generate, and extract insights from unstructured clinical text, supporting decision-making and enabling richer cross-domain knowledge integration\cite{singhal2023large, lee2023biogpt}. Given the abundance of textual data—ranging from case reports to clinical notes—LLMs hold unique promise for rare diseases.

However, the application of LLMs in this domain remains fragmented and insufficiently characterized. Comprehensive evaluation is lacking, with open questions surrounding their effective deployment, potential benefits, and methodological limitations in this sensitive, data-scarce context. A systematic survey is therefore critical to map current approaches, highlight successful applications, and identify challenges, thereby guiding clinicians, researchers, and practitioners toward informed and responsible use\cite{he2023survey, zhao2023survey,mercatelli2021exploiting}.

This review aims to provide that comprehensive synthesis, detailing how LLMs are currently leveraged to improve diagnosis, clarify disease mechanisms, and inform therapeutic strategies in rare diseases. We also explore emerging efforts to integrate LLMs with multimodal datasets—including genomic sequences, medical imaging, and electronic health records—underscoring the potential to advance personalized medicine and improve patient outcomes. By synthesizing recent advances and outlining key future directions, we chart a path forward for harnessing LLMs to meet the unique challenges of rare disease research and care.

\subsection{Overview of Rare Diseases and Diagnostic Challenges}
\label{sec:rare_diseases}

Rare diseases are defined as medical conditions that affect a small percentage of the population but have a significant clinical and social impact on patients and their families~\cite{rodwell2015rare}. In the European Union (EU), a disease is considered rare if it affects fewer than 5 individuals per 10,000~\cite{EU_RareDiseases}. Despite their low individual prevalence, rare diseases collectively affect between 27 and 36 million people in the EU alone. Approximately 80\% of these diseases are genetic, and 70\% manifest during childhood.

In the United States, the Orphan Drug Act defines a rare disease as one affecting fewer than 200,000 people~\cite{FDA_RareDiseases}. This definition underpins eligibility for incentives and regulatory support mechanisms managed by the Food and Drug Administration (FDA), which facilitates the development of therapies for underserved patient populations.

Similarly, in China, it is estimated that over 20 million people are affected by rare conditions~\cite{tang2023release}. The 2023 release of the second official list of rare diseases brought the total number of recognized conditions to 207, reflecting increasing national commitment to this public health issue.

Globally, rare diseases affect between 3.5\% and 5.9\% of the population~\cite{NguengangWakap2020}. These disorders range from mild to severe and are often chronic, progressive, and life-threatening. Common examples include amyloidosis~\cite{mallus2023treatment}, cystic fibrosis~\cite{ong2023cystic}, Huntington’s disease~\cite{stoker2022huntington}, and Duchenne muscular dystrophy~\cite{bez2023duchenne}.

Despite their diversity, rare diseases share common systemic challenges. These include:

\begin{itemize}
    \item Limited patient populations, which hinder large-scale studies and clinical trials.
    \item Genetic heterogeneity, with many conditions caused by rare or private mutations.
    \item Lack of high-quality clinical and genomic data.
    \item Delayed diagnosis due to nonspecific symptoms.
\end{itemize}

A major obstacle to progress in rare disease research is the scarcity of structured and interoperable clinical data. As highlighted by Hageman \textit{et al.}~\cite{hageman2023systematic}, disease registries are essential for enabling research but often face challenges related to standardization, long-term funding, and data sharing regulations. Moreover, few registries incorporate multi-omics data, limiting the ability to establish robust genotype-phenotype correlations.

Therapeutic development for rare diseases also faces substantial hurdles. According to Chen \textit{et al.}~\cite{chen2024trends}, although technologies such as CRISPR/Cas9 and RNA-based treatments have shown promise, the cost of developing orphan drugs remains high, and the number of approved products remains limited. National healthcare systems may find it difficult to justify the high prices of such therapies, despite their clinical necessity.

Another persistent issue is the diagnostic delay experienced by many patients. The so-called “diagnostic odyssey”~\cite{visibelli2023impact} may last several years and involve multiple misdiagnoses. For instance, hereditary transthyretin amyloidosis (hATTR) is often mistaken for common conditions like heart failure or neuropathy~\cite{mallus2023treatment}, leading to delayed intervention and irreversible organ damage.

\begin{figure}[ht]
\centering
\includegraphics[width=0.9\linewidth]{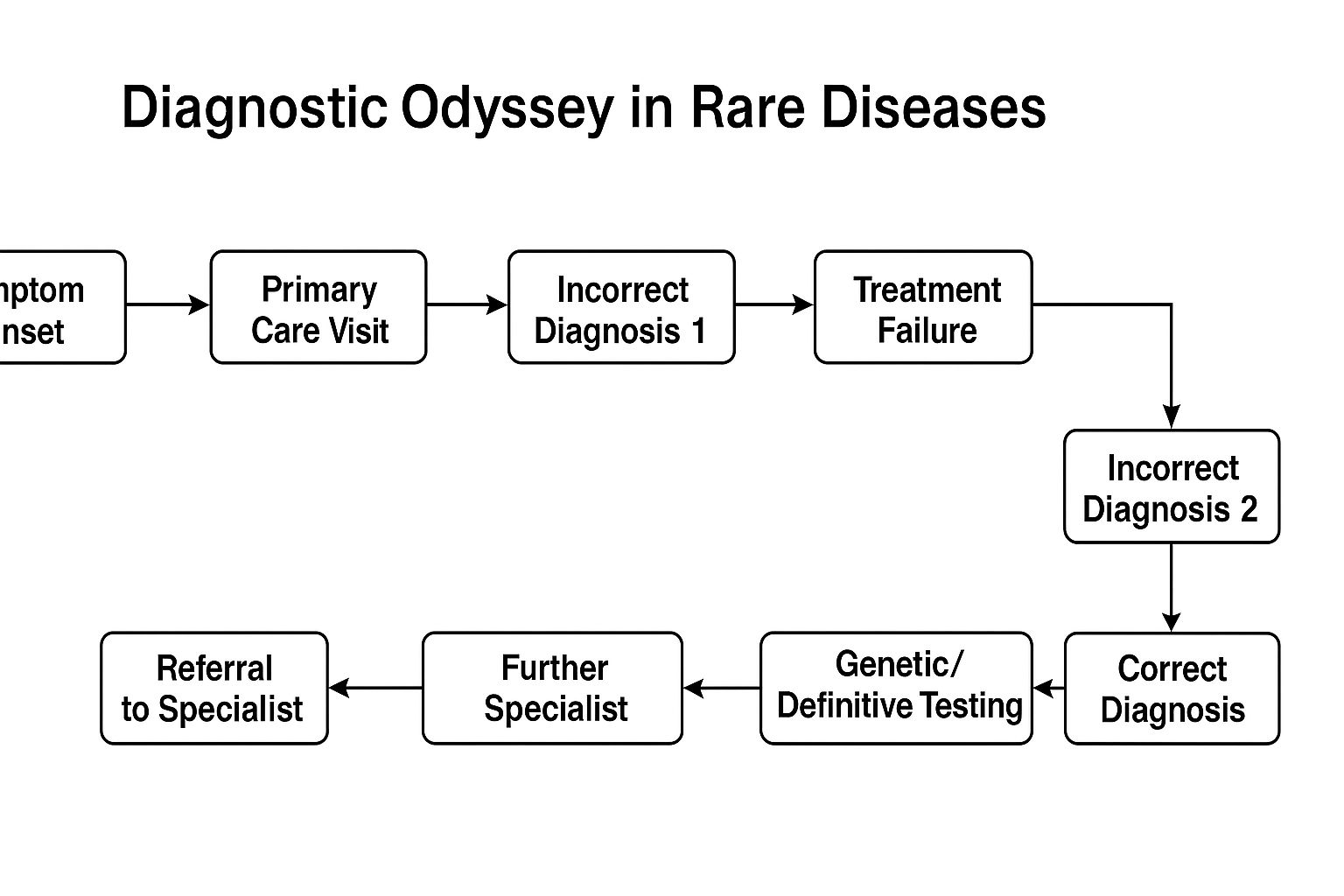}
\caption{Flowchart of the diagnostic journey for rare disease patients. The path often includes multiple incorrect diagnoses, specialist referrals, and redundant testing before the correct diagnosis is reached, often several years after symptom onset.}
\label{fig:diagnostic_odyssey}
\end{figure}

Fig.~\ref{fig:diagnostic_odyssey} presents a schematic of the diagnostic journey typically encountered by patients with rare diseases.

\section{Methodology Selection of Papers}

\begin{figure}[H]
    \centering
    \includegraphics[width=\textwidth]{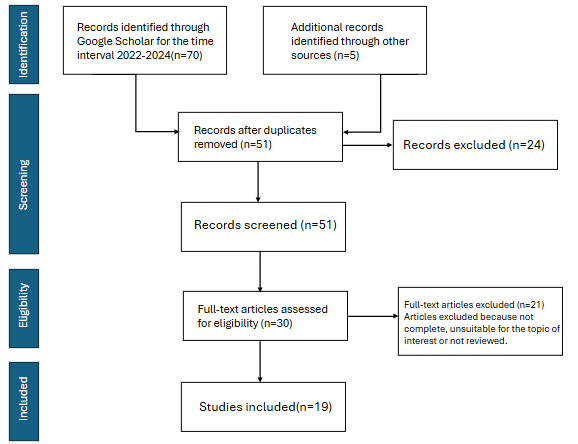}
    \caption{PRISMA flow diagram illustrating the systematic review process. Searches included combinations of "LLM", "Rare Diseases", "Data Analysis", and "Questionnaires" - (2022-2024).}
    \label{fig:prisma-diagram}
\end{figure}

The following section describes the different methodologies adopted to identify and analyze relevant scientific studies on the application of traditional Large Language Models (LLM) in the context of rare diseases. A systematic review of the literature was conducted following the PRISMA (Preferred Reporting Items for Systematic Reviews and Meta-Analyses) guidelines, which offer a structured approach for the identification, selection and inclusion of scientific literature. As shown in Figure \ref{fig:prisma-diagram}, the PRISMA process was articulated in four phases:
\begin{itemize}

\item \textbf{Identification}: A comprehensive search was initially performed mainly through Google Scholar, considering articles published between 2022 and 2024, with keywords such as "LLM", "Rare Disease", "Questionnaires" and "Data Analysis". This search identified 70 potentially relevant articles. Another 5 studies were identified through other sources, such as reference lists and preprint archives, bringing the total to 75.
\item \textbf{Screening}: After the removal of 24 duplicate or clearly irrelevant entries (based on titles and abstracts), 51 records were retained for screening.

\item \textbf{Eligibility}: These 51 studies were subsequently analyzed in detail. Of these, 30 met the initial inclusion criteria and were assessed by full-text review. During this phase, 21 articles were excluded due to incompleteness, lack of peer review, or insufficient relevance to the topic of LLMs in rare diseases.

\item  \textbf{Included}: A total of 19 studies were ultimately included in the final analysis.

\end{itemize}

Below is an in-depth description of the articles selected and, in addition, it is important to highlight that the analysis of the selected literature revealed two distinct methodological approaches regarding the application of LLMs in the context of rare diseases. Some studies, in fact, are based on the use of questionnaires, mainly used to evaluate the perceived accuracy, reliability or usefulness of the model results. Other studies, instead, adopt a more computational approach, aiming to explore the ability of LLMs to extract, synthesize or generate information related to rare diseases. These works evaluate the models based on their ability to perform tasks such as identifying rare conditions in unstructured texts, supporting differential diagnosis or retrieving clinically relevant knowledge from medical corpora.

Table \ref{tab:article_summary} presents a summary of the main articles analyzed that focus on specific rare diseases (n=4). Each column of the table has been designed to highlight methodological elements relevant for the analysis.
The “\textbf{Article}” column provides the full title of each included study, while “\textbf{Studied Disease}” indicates the rare disease being investigated, such as glaucoma, sarcoidosis, Kinsbourne disease, or amyloidosis. The “\textbf{Objective}” column follows, summarizing the main purpose of the study, which can range from evaluating diagnostic accuracy to the ability to extract information from textual data. The “\textbf{Type of Study}” field distinguishes between monomodal and multimodal studies; in this selection all studies are monomodal, that is, based on a single type of input. The “\textbf{Input Data}” column then specifies the nature of the data used, which can be structured questionnaires or unstructured texts (such as posts from social media or clinical forums).

Then there is “\textbf{Access Type}”, which indicates whether the model used in the study is closed-source or not, and finally “\textbf{Pipeline}”, which indicates whether the model was used in stand-alone mode or integrated into a larger system. In all the articles analyzed, the models were tested in closed environments and independently, without clinical pipelines or external tools to support them.
\\

\begin{table}[H]
\centering
\small
\begin{adjustbox}{width=\textwidth}
\begin{tabular}{|>{\raggedright\arraybackslash}p{4cm}|
                >{\raggedright\arraybackslash}p{2.3cm}|
                >{\raggedright\arraybackslash}p{3.5cm}|
                >{\centering\arraybackslash}p{1.8cm}|
                >{\raggedright\arraybackslash}p{2.5cm}|
                >{\centering\arraybackslash}p{1.8cm}|
                >{\centering\arraybackslash}p{1.8cm}|}
\hline
\textbf{Article} & \textbf{Studied Disease} & \textbf{Objective} & \textbf{Type of Study} & \textbf{Input Data} & \textbf{Access Type} & \textbf{Pipeline} \\
\hline
Assessment of a Large Language Model’s Responses to Questions and Cases About Glaucoma and Retina Management \cite{Huang2024} & Glaucoma & Evaluation of diagnostic accuracy and completeness & Monomodal & Questionnaire & Closed & Standalone \\
\hline
Understanding Sarcoidosis Using Large Language Models and Social Media Data \cite{MilesXi2024} & Sarcoidosis & Extracting insight from patient Reddit data & Monomodal & Unstructured textual data & Closed & Standalone \\
\hline
High accuracy but limited readability of large language model-generated responses to frequently asked questions about Kienböck’s disease \cite{Asfuroglu2024} & Kienböck’s disease & General knowledge, diagnosis and treatment & Monomodal & Questionnaire & Closed & Standalone \\
\hline
A multidisciplinary assessment of ChatGPT’s knowledge of amyloidosis \cite{RyanC2024}& Amyloidosis & Clinical question-answering performance evaluation & Monomodal & Questionnaire & Closed & Standalone \\
\hline
\end{tabular}
\end{adjustbox}
\caption{Summary of articles on specific rare disease-related LLM evaluations}
\label{tab:article_summary}
\end{table}

\section{Research in LLM and Rare Disease }

Research into the use of Large Language Models (LLMs) in the context of rare diseases is gaining momentum, but the field remains highly diverse. The variety in research goals, data sources, technical implementations, and clinical applications makes it difficult to directly compare studies. To address this issue, we propose a structured classification framework that can systematically analyze and interpret the literature.

This framework, detailed in Table 1, describes several key dimensions for categorizing LLM-based studies on rare diseases. Our aim is to provide a clearer understanding of the research landscape, highlight emerging trends, and identify areas that require further exploration or standardization.

Standardization encompasses several fundamental aspects. First, we consider the type of disease focus: whether the study targets a single rare disease, a class of disease (e.g. genetic disorders, ophthalmologic diseases), or adopts a general-purpose approach applicable to multiple rare conditions.  Andrew et al., \cite{andrew2024evaluating}, present the evaluation of the performances of the LLMs in the diagnosis of pediatric rare diseases. Asfuroglu et al. \cite{Asfuroglu2024} apply LLM to diagnose a rare muscoloskeletal disease; Huang et al. \cite{Huang2024} apply LLM to the study of glaucoma. Finally,  most approaches focus on the whole spectrum of rare diseases, and limitations are often related to the availability of specific datasets \cite{hybrid_framework,Shyr2024,yang2024rdguru,kim2024assessing,abdullahi2024learning,zampatti2024innovations,young2025diagnostic,rarebench,soman2024zebra,MilesXi2024,cao2024automatic,oniani2023large,do2024assessing}.

Second, we categorize the study's objectives, which may include diagnosis, prognosis, or research-oriented tasks such as gene prioritization and data extraction.

Large Language Models (LLMs) can be effectively adapted for the diagnosis of rare diseases by using a combination of techniques: prompting, retrieval-augmented generation (RAG), fine-tuning, and domain-specific pre-training \cite{ong2025scoping}
. 

Prompting techniques, such as role-based prompts and chain-of-thought instructions, help guide LLMs to systematically analyze patient phenotypes and suggest potential differential diagnoses. Dynamic few-shot prompting enhances model performance by providing carefully selected clinical examples within the input, which improves the model's ability to generalize to rare and complex cases.

Retrieval-augmented generation further supports prompting by supplying LLMs with curated external medical knowledge at query time. This enriches the model’s contextual understanding and enhances the accuracy of diagnostic suggestions.

In addition to input-level adaptations, fine-tuning and pre-training approaches contribute to deeper model specialization. Fine-tuning involves updating model parameters using rare disease-specific corpora, allowing LLMs to internalize patterns of symptom-disease associations and clinical reasoning processes. Domain-specific pre-training, which exposes models to extensive biomedical literature, structured phenotype ontologies, and curated case reports from the early stages of training, establishes a foundational understanding of rare disease contexts.

Large language models (LLMs) are being used to assist healthcare professionals by improving the precision and timeliness of diagnostic processes; they are capable of synthesizing vast amounts of clinical data—including patient history, imaging reports, and laboratory findings—to recognize disease patterns, reduce diagnostic errors, and even flag conditions at an early stage, which is particularly valuable for diseases with subtle or non-specific symptoms \cite{}.

When applied to prognosis, LLMs support the ongoing evaluation of a patient's condition over time by analyzing longitudinal health records, identifying trends or risk factors indicative of disease progression, and generating predictive models that inform clinicians about likely future developments; this enables more proactive management strategies, timely interventions, and the personalization of follow-up plans based on individual risk profiles \cite{Bergman2011}.

In research-focused applications, LLMs serve as powerful tools for uncovering complex genotype-phenotype associations by parsing large-scale genomic and clinical datasets, identifying meaningful correlations, and suggesting candidate genes for further study; additionally, these models can be employed to construct interactive systems that comprehend and respond to specialized biomedical questionnaires, facilitating automated data collection, hypothesis generation, and support for literature-based discovery in biomedical sciences \cite{chung2024large}.

In addition, we classify the input data modalities used, ranging from free-text clinical reports to structured questionnaires and laboratory results. The framework also captures the type of LLM employed, including general-purpose models such as GPT-4 and domain-specific fine-tuned models. Importantly, we note the availability and accessibility of these models, distinguishing between open-source, commercial, closed-source, and freely accessible non-open platforms.

Finally, we assess the role of the LLM within the broader computational pipeline, determining whether it acts as a standalone diagnostic tool or is integrated within a multicomponent system involving other data processing or inference modules. This comprehensive classification allows for a richer understanding of how LLMs are being adapted to the rare disease domain and provides a foundation for more consistent benchmarking and evaluation in the future. Research into the use of Large Language Models (LLMs) in the context of rare diseases is gaining momentum, but the field remains highly diverse. The variety in research goals, data sources, technical implementations, and clinical applications makes it difficult to make direct comparisons between studies. To address this issue, we propose a structured classification framework that can systematically analyze and interpret the literature.

This framework, detailed in Table \ref{tab:research}, outlines several key dimensions to categorize LLM-based studies on rare diseases. Our aim is to provide a clearer understanding of the research landscape, highlight emerging trends, and identify areas that require further exploration or standardization.

Standardization encompasses several fundamental aspects. First, we consider the type of disease focus: whether the study targets a single rare disease, a class of disease (e.g. genetic disorders, ophthalmologic diseases), or adopts a general-purpose approach applicable to multiple rare conditions. Second, we categorize the study's objectives, which may include diagnosis, prognosis, or research-oriented tasks such as gene prioritization and data extraction. 

In addition, we classify the input data modalities used, ranging from free-text clinical reports to structured questionnaires and laboratory results. The framework also captures the type of LLM employed, including general-purpose models such as GPT-4 and domain-specific fine-tuned models. Importantly, we note the availability and accessibility of these models, distinguishing between open-source, commercial, closed-source, and freely accessible non-open platforms.

Finally, we assess the role of the LLM within the broader computational pipeline, determining whether it acts as a standalone diagnostic tool or is integrated within a multi-component system involving other data processing or inference modules. This comprehensive classification allows for a richer understanding of how LLMs are being adapted to the rare disease domain and provides a foundation for more consistent benchmarking and evaluation in the future.

\subsubsection{Types of Input Data}

The selected studies involving specific rare diseases use different types of input data, ranging from structured clinical questionnaires to unstructured online content. For example, Huang et al. \cite{Huang2024} used 20 expert-generated clinical questions—10 each on glaucoma and 10 on retinal diseases—randomly selected from the "Ask an Eye Doctor" section of the American Academy of Ophthalmology website, along with 20 anonymized patient cases from ophthalmology clinics. These were used to assess the accuracy of GPT-4 responses for glaucoma and retinal diseases. \\
In contrast, Xi et al. \cite{MilesXi2024} analyzed user-generated content on Reddit, mining sarcoidosis-related posts to explore patients’ self-reported symptoms and their assessments of medications, diagnoses, and experiences. Asfuroğlu et al. \cite{Asfuroglu2024} developed a frequently asked questions (FAQ) questionnaire on Kienböck's disease. The terms "Kienböck's disease" and "lunar avascular necrosis" were used to collect the most commonly searched terms from the three major search engines: Google, Bing, and Yahoo; then, relevant FAQs were identified from reliable medical websites. Finally, King et al. \cite{RyanC2024} evaluated ChatGPT's ability to answer a total of 98 amyloidosis-related questions from medical societies and institutions. To provide a more comprehensive perspective for patients, questions from amyloidosis Facebook support groups were also included. Of the total questions, 56 were general amyloidosis topics, while 42 focused on specific subfields, including cardiology and neurology.

\subsubsection{Prompting Strategies}

In the reviewed studies on the use of LLM in the field of rare diseases, in terms of prompting strategies, most of them used simple prompts, usually zero-shot, often instructing the model to emulate a healthcare professional. For example, instruction-based prompts were used with explicit directions for GPT-4 to respond in the style of a clinical note, including common abbreviations in ophthalmology \cite{Huang2024}.



\section{Dataset}

This section outlines key datasets referenced in the reviewed studies, emphasizing their core characteristics and accessibility. The selected literature draws on a broad spectrum of data sources, which can be grouped into four main categories: clinical datasets, biomedical ontologies and knowledge bases, textual corpora and bibliographic repositories, and patient-derived content from social media platforms. Each data type serves distinct purposes—ranging from automated diagnostic support to medical language understanding and generation—highlighting the multifaceted role of data in advancing biomedical natural language processing.
.

\subsection{Clinical Datasets}

Clinical datasets are central to biomedical research, providing the foundation for developing and validating NLP systems tailored to medical applications. Typically derived from electronic health records (EHRs), clinical notes, laboratory findings, and demographic data, these resources capture the complexity of real-world healthcare settings. They support predictive modelling, enable automated phenotyping, and power clinical decision support tools. Among the datasets most frequently cited in the reviewed studies are MIMIC-III, MIMIC-IV, RareDis, RAMEDIS, and PUMCH (Table~\ref{tab:source}).

\begin{table}[H]
\centering
\begin{tabular}{|l|l|l|}
\hline
\textbf{Source} & \textbf{Typology} & \textbf{Material} \\
\hline
Ramedis & Open & Rare disease data \& clinical cases \\
MIMIC-III & Open (with registration) & ICU patient data (clinical, demographics) \\
MIMIC-IV & Open (with registration) & ICU patient data (clinical, time-series) \\
RareDis & Open & Rare disease phenotypic data \\
PUMCH & Restricted (institutional access) & Patient records (clinical \& genomic) \\
\hline
\end{tabular}
\caption{Clinical Dataset}
\label{tab:source}
\end{table}

\textbf{MIMIC-III} (Medical Information Mart for Intensive Care III)~\cite{johnson2016mimic} is one of the most widely used clinical datasets in medical NLP. Developed by the MIT Lab for Computational Physiology, it contains de-identified health records from over 40,000 patients admitted to Beth Israel Deaconess Medical Center between 2001 and 2012. The dataset includes both structured data—such as laboratory test results, medications, demographics, and vital signs—and unstructured clinical notes. MIMIC-III has been extensively adopted for patient phenotyping, disease classification, and predictive modelling \cite{guzzi2020biological}.

\textbf{MIMIC-IV}~\cite{johnson2023mimic} is the successor to MIMIC-III and extends its coverage to the years 2008–2019. It introduces refined data structuring, improved linkage between hospital departments, and enhanced consistency across patient records. Like its predecessor, it includes both structured and unstructured components and continues to serve as a cornerstone for reproducible research in medical artificial intelligence and clinical NLP \cite{hiram2022disease}.

Another frequently cited dataset is \textbf{RareDis}~\cite{martinez2022raredis}, a resource specifically designed to support rare disease research. It consists of annotated clinical case descriptions sourced from the scientific literature and expert-curated repositories. Each case typically includes a narrative of symptoms and diagnostic conclusions, often complemented by ontological annotations based on systems such as the Human Phenotype Ontology (HPO) or Orphanet. RareDis exists in multiple versions, commonly identified by year (e.g., RareDis 2022, RareDis 2023), which vary in format, language, and coverage. This adaptability has enabled its application across a range of NLP tasks, including few-shot learning and disease classification.

\textbf{RAMEDIS} (Rare Metabolic Diseases Information System)~\cite{topel2006ramedis} is a specialized database designed to support diagnosis and research in the field of rare metabolic disorders. It integrates clinical, biochemical, and genetic data and offers multilingual access. Developed in the early 2000s, RAMEDIS is aligned with standardized disease classifications and has proven useful for both clinical decision support and interoperability in biomedical NLP applications.

Several studies also reference the so-called \textbf{PUMCH dataset}, derived from clinical records at the Peking Union Medical College Hospital. While not a standardized or publicly available dataset, it comprises institutional EHRs, diagnostic notes, and genomic data used internally for developing and validating AI models. Despite its restricted access, PUMCH has contributed significantly to rare disease informatics, particularly within studies focused on Chinese populations.

\subsection{Biomedical Ontologies and Knowledge Bases}

Biomedical ontologies and knowledge bases offer structured representations of complex medical and biological entities, enabling semantic interoperability, data harmonization, and automated reasoning. These resources play a pivotal role in augmenting clinical datasets, supporting entity recognition, and enhancing the transparency and interpretability of machine learning models. In the reviewed literature, ontologies are frequently employed to standardize disease terminology, map genetic and phenotypic relationships, and enable a range of downstream NLP applications. The main ontologies and knowledge bases identified across studies are summarized in Table~\ref{tab:ontology}.

\begin{table}[H]
\centering
\begin{tabular}{|l|l|l|}
\hline
\textbf{Ontology} & \textbf{Typology} & \textbf{Material} \\
\hline
ORDO & Open & Rare diseases classification \\
ORPHANET & Open & Rare diseases database \\
OMIM & Open (partial access) & Genetic disorders \& phenotypes \\
MONDO & Open & Disease ontology \\
HPO & Open & Phenotype descriptions \\
UMLS & Restricted (free access with registration) & Medical vocabularies \& ontologies \\
\hline
\end{tabular}
\caption{Ontology Table}
\label{tab:ontology}
\end{table}
\textbf{ORDO} (Orphanet Rare Disease Ontology)~\cite{vasant2014ordo} provides a structured representation of the rare disease domain, capturing relationships between disorders, phenotypes, inheritance patterns, and associated genes. Developed in OWL (Web Ontology Language), ORDO is designed for semantic reasoning and classification and is frequently used to annotate patient records in rare disease registries, linking phenotypic and molecular data. Its machine-readable structure supports both clinical interpretation and automated disease matching in NLP pipelines.

\textbf{Orphanet}~\cite{nguengang2020estimating} is a comprehensive, curated knowledge base maintained by a multinational consortium coordinated by INSERM. It encompasses more than 6,000 rare diseases, offering detailed metadata on prevalence, diagnostic criteria, associated genes, clinical features, and therapeutic options. In biomedical NLP, Orphanet is widely used for lexicon development, annotation tasks, and extraction of structured knowledge from unstructured medical narratives. Its standardized nomenclature, including ORPHA codes, enhances data interoperability across studies.

\textbf{OMIM} (Online Mendelian Inheritance in Man)~\cite{mckusick2007mendelian} is a foundational reference catalog of human genes and genetic disorders, currently comprising over 15,000 entries. It includes curated descriptions of gene-disease associations, inheritance patterns, mutation profiles, and phenotypic variability. In computational research, OMIM is frequently used for gene–disease mapping and serves as a benchmark for genetic entity recognition and linkage tasks in biomedical NLP.

Three additional ontologies frequently cited in the literature are \textbf{MONDO}, \textbf{HPO}, and \textbf{UMLS}.

\textbf{HPO} (Human Phenotype Ontology)~\cite{kohler2021human} comprises over 13,000 phenotypic terms organized in a directed acyclic graph, capturing hierarchical relations among clinical features. First introduced in 2008, it is widely used in clinical genomics for differential diagnosis and is integral to many NLP tasks involving phenotype extraction and classification in rare disease contexts.

\textbf{MONDO}~\cite{vasilevsky2020mondo} aims to unify disease terminologies from diverse ontological sources, including OMIM, Orphanet, ICD, and SNOMED CT. Through logical mapping and hierarchical alignment, MONDO supports consistent annotation across heterogeneous corpora and enables semantic harmonization of disease mentions in NLP models.

\textbf{UMLS} (Unified Medical Language System)~\cite{bodenreider2004unified}, developed by the U.S. National Library of Medicine, integrates over 200 biomedical vocabularies and ontologies. It provides tools for synonym mapping, semantic typing, and multilingual alignment. Although not an ontology in the strict sense, UMLS is essential for concept normalization and semantic integration in clinical NLP systems and remains a backbone for many large-scale medical knowledge extraction frameworks.

\subsection{Textual Corpora and Bibliographic Sources}
Text corpora and bibliographic resources, summarized in Table~\ref{tab:pubmed}, play a crucial role in the development and evaluation of NLP systems aimed at understanding medical language, generating clinical summaries, and responding to domain-specific queries. These corpora—comprising biomedical publications, clinical guidelines, and curated question–answer pairs—provide rich, high-quality text that reflects the linguistic and conceptual complexity of medical discourse. In the reviewed studies, such resources are widely used to fine-tune large language models and assess their performance on tasks aligned with real-world clinical applications.

\begin{table}[H]
\centering
\begin{tabular}{|l|l|l|}
\hline
\textbf{Source} & \textbf{Typology} & \textbf{Material} \\
\hline
PubMed & Open & Abstracts \\
PubMedQA & Open & QA pairs, abstracts \\
MedQA & Restricted / Open & Exam questions \\
\hline
\end{tabular}
\caption{PubMed, PubMedQA, MedQA Table}
\label{tab:pubmed}
\end{table}
\textbf{PubMed}~\cite{white2020pubmed} is among the most extensively used resources in biomedical NLP. Maintained by the National Center for Biotechnology Information (NCBI), it provides access to millions of peer-reviewed scientific articles. PubMed abstracts serve as a foundational corpus for training domain-specific language models and are frequently used in information retrieval and question answering tasks.

\textbf{PubMedQA}~\cite{jin2019pubmedqa} is a curated dataset designed for biomedical question answering. It consists of clinical yes/no/maybe questions paired with corresponding PubMed abstracts. This resource is commonly used to assess the reasoning capabilities and factual consistency of NLP models in processing medical literature.

\textbf{MedQA}~\cite{jin2021disease} comprises multiple-choice questions derived from official medical licensing examinations. The dataset presents complex, clinically relevant scenarios and has become a benchmark for evaluating the diagnostic reasoning performance of large language models, including GPT-based systems, in tasks related to medical education and clinical decision-making.

\subsection{Data from Social Networks}

Social networks have now emerged as highly supportive, yet complex, sources of data for medical research. Unlike curated clinical datasets or structured ontologies, social media platforms provide access to patient interactions and expressions of symptoms and experiences of those suffering from specific diseases. This content presents both opportunities and challenges: while rich in context and factual details, it is also unstructured and requires more specialized natural language processing techniques to extract clinically relevant information. Among these platforms, \textbf{Reddit} has gained particular attention due to its thematic organization and the presence of active health-related communities. \\
It is \cite{proferes2021studying} a social network structured around various topics-focused forums known as "subreddits," where users can share experiences, ask questions, and discuss topics anonymously. Its semi-structured format of posts and comments, and the presence of topic-specific communities make it an attractive environment for mining patient-reported information.
In recent years, Reddit has been used in medical NLP studies to explore awareness of rare diseases, mental health trends, and drug side effects. Researchers are using Reddit to collect large volumes of real-world text data, which can then be annotated, clustered, or modeled to identify emerging health issues or train LLMs in understanding common language.

\section{Challenges}

Table~\ref{tab:limitations} outlines the main limitations associated with using large language models (LLMs) for supporting rare disease diagnosis. These limitations can be broadly categorized into technical, clinical, and governance-related challenges.

First, LLMs are inherently prone to hallucinations, which means they can generate outputs that are linguistically coherent but factually incorrect or fabricated. This issue is particularly concerning in the context of rare diseases, where the scarcity of reliable training examples increases the risk of model errors. Due to the underrepresentation of rare disease data in general corpora, LLMs may produce inaccurate associations between clinical features and specific conditions.

Zero-shot prompting, where models are queried without domain-specific fine-tuning, faces unique challenges due to the complexity and specificity required for rare disease diagnosis. Generalization failures are common, as LLMs have limited capacity to extrapolate reliably to medical domains for which they have not been explicitly trained. Additionally, the lack of curated, high-quality datasets for rare diseases restricts both the models' initial learning and opportunities for external validation.

General-purpose LLMs are mainly trained on non-peer-reviewed, heterogeneous internet sources, which diminishes their reliability for clinical application. Consequently, the outputs often deviate from established evidence-based diagnostic standards, posing potential patient safety risks. While specialized medical LLMs show improved performance, they still suffer from insufficient coverage of rare diseases due to the limited availability of appropriate training materials.

Fine-tuning on small rare disease datasets carries the risk of overfitting, where the model may capture dataset-specific noise rather than generalizable patterns. This overfitting can amplify biases present in the training data, such as demographic underrepresentation, leading to inequities in diagnostic outcomes.

LLMs also struggle to recognize subtle clinical nuances that are often critical for identifying rare diseases. Specialized models require ongoing updating and curation to remain relevant as medical knowledge evolves. However, few studies have systematically validated LLM performance specifically for rare disease diagnostic applications, which limits confidence in their deployment.

Another significant barrier is the opacity of LLMs, as their internal decision-making processes are largely uninterpretable. This lack of explainability complicates their integration into clinical workflows and undermines clinician trust. Moreover, errors in model outputs may propagate through diagnostic chains, compounding clinical risks.

Current LLMs often fail to adequately capture demographic variability in the presentation of rare diseases, further exacerbating the risks of diagnostic disparities. Additionally, regulatory and ethical challenges persist, particularly concerning the deployment of unvalidated diagnostic tools in clinical settings.

Finally, while specialized medical LLMs demonstrate superior performance compared to general-purpose models, both types require rigorous validation, governance, and continuous oversight. Without these safeguards, the deployment of LLMs could perpetuate and potentially worsen diagnostic inaccuracies and disparities, especially among vulnerable populations affected by rare diseases.

In conclusion, careful design, domain-specific adaptation, external validation, and human-centered oversight are essential for the safe and effective integration of LLMs into rare disease diagnostic workflows.

\begin{table}[ht]
\centering
\caption{Limitations of Large Language Models (LLMs) in Supporting Rare Disease Diagnosis}
\begin{tabularx}{\textwidth}{|p{4cm}|X|}
\hline
\textbf{Limitation} & \textbf{Description} \\
\hline
Hallucination Risk & LLMs may produce plausible but factually inaccurate or fabricated outputs, increasing diagnostic risk. \\
\hline
Data Scarcity & Rare diseases are underrepresented in training data, limiting model exposure and learning. \\
\hline
Zero-Shot Prompting Challenges & Lack of domain-specific adaptation hampers model ability to handle rare disease cases effectively. \\
\hline
Limited Generalization & LLMs generalize poorly to highly specialized or unfamiliar medical scenarios. \\
\hline
Training Data Quality & Scarcity of curated datasets restricts model performance and validation opportunities. \\
\hline
Source Quality Limitations & General LLMs are trained on non-peer-reviewed sources, reducing clinical reliability. \\
\hline
Deviation from Medical Standards & Model outputs may not align with evidence-based diagnostic protocols. \\
\hline
Constraints of Specialized LLMs & Even fine-tuned models show limited rare disease representation. \\
\hline
Risk of Overfitting & Fine-tuning on small datasets can reduce model generalizability. \\
\hline
Amplification of Biases & Existing biases in clinical data may be reinforced in model outputs. \\
\hline
Misinterpretation of Clinical Nuances & Subtle symptomatology critical for rare diseases may be overlooked. \\
\hline
Need for Continuous Updating & Specialized models require ongoing curation to remain clinically relevant. \\
\hline
Lack of Systematic Validation & Few studies assess LLM performance specifically for rare disease diagnosis. \\
\hline
Opacity and Limited Explainability & LLMs’ black-box nature complicates clinical decision-making and trust. \\
\hline
Error Propagation Risk & Errors in LLM outputs can cascade through diagnostic workflows. \\
\hline
Demographic Variability & LLMs insufficiently capture diverse disease presentations across populations. \\
\hline
Regulatory and Ethical Challenges & Unvalidated outputs pose risks for ethical clinical deployment and approval. \\
\hline
Performance Gap Between General and Specialized LLMs & General models perform markedly worse than domain-specific models, but both require scrutiny. \\
\hline
Risk of Diagnostic Disparities & Improper deployment could exacerbate inequities in rare disease diagnosis. \\
\hline
Need for Governance and Oversight & Human supervision and rigorous frameworks are essential for safe use. \\
\hline
\end{tabularx}
\label{tab:limitations}
\end{table}


\section{Future Perspectives}

Rare disease research faces a significant data scarcity challenge that limits the development of diagnostic tools, therapeutic strategies, and essential biological insights. In this context, Large Language Models (LLMs) offer a transformative opportunity for data augmentation. Future research should focus on systematically utilizing LLMs to simulate, expand, and enrich rare disease datasets with synthetic yet clinically plausible information \cite{mitani2020small}.

Firstly, LLMs could be fine-tuned to generate synthetic patient cases that include structured phenotypic profiles, detailed clinical narratives, and plausible disease progression trajectories. Such augmented datasets could enhance the training of downstream machine learning models, particularly in scenarios where real-world case numbers are prohibitively small \cite{rarebench}. The RareBench initiative \cite{rarebench} already highlights the difficulties LLMs encounter when faced with limited, heterogeneous clinical examples and points toward dynamic few-shot prompting as a promising strategy for improving data diversity.

Secondly, retrieval-augmented generation (RAG) frameworks can be integrated to ensure that LLM-generated content is grounded in verified medical knowledge, thereby minimizing the risk of hallucinations. Future LLM-based augmentation systems must closely couple generation with retrieval from curated rare disease knowledge bases, dynamically constraining outputs to maintain clinical accuracy.

Third, LLMs have significant potential to expand phenotype ontologies and fill gaps in rare disease knowledge graphs. LLMs could propose novel associations of symptoms, patterns of comorbidity, or disease pathways that could otherwise remain underreported in the clinical literature. These hypotheses could then be validated experimentally, accelerating the pace of knowledge discovery.

Fourth, adaptive domain-specific pre-training strategies are essential. Rather than relying solely on generalist models, future research should explore lightweight, continual pre-training of LLMs on rare disease–focused corpora, including clinical reports, case studies, and genetic databases. This would enhance the biological and clinical relevance of synthetic outputs without necessitating a complete model retraining.

However, realizing this vision requires careful attention to validation and ethical considerations. Synthetic datasets must be rigorously evaluated for realism, diversity, and potential bias propagation. Furthermore, governance frameworks should be developed to delineate appropriate use cases for synthetic data, ensuring that augmented datasets do not inadvertently reinforce diagnostic inaccuracies or demographic disparities.

Finally, collaborations between clinical experts, data scientists, and AI ethicists will be crucial in co-developing standards for the responsible use of LLMs in rare disease research. Establishing transparent benchmarks for evaluating synthetic data quality, clinical plausibility, and downstream impact will be a priority.

With careful design, grounding, and oversight, LLM-driven data augmentation could substantially mitigate current data limitations, accelerate model development, and ultimately improve outcomes for patients affected by rare diseases.

\section{Conclusion}
Even while big language models have advanced a lot, there is still much work to be done in terms of integrating multimodal data for rare diseases. Although rare diseases necessitate a more comprehensive approach, the majority of current research on LLM applications is unimodal, such as in natural language processing \cite{denecke2024} or medical imaging \cite{hasani2022}. These diseases require a technology that can combine information from genomic data, imaging investigations, laboratory results and textual descriptions. Such integration now represents an unmet need for LLMs' full potential to offer thorough insights for diagnosis and therapy. 
Our experimentation with different LLMs, however, showed promising results regarding their potential to assist in diagnosis. The application of LLMs in simulating interactions with patients and analysing medical texts has shown the ability to identify patterns and correlations that can help healthcare professionals. 

Therefore, further development in this area is necessary to improve LLM's capacity to handle multimodal data. The diagnosis and treatment of rare diseases necessitate the integration of text, photos, structured data and even longitudinal patient histories, even if the existing models are excellent at text-based information processing. Multimodal AI frameworks, according to Ou et al. \cite{ou2024}, have the potential to completely transform diagnostics by revealing intricate connections between various data sources that could otherwise go overlooked. 

For instance, the early and more precise identification of rare illnesses may be made possible by combining phenotypic descriptions with genetic variants and imaging biomarkers. The lack of sizable, harmonized multimodal datasets and the processing power needed to train such models are two of the many challenges that impede this integration. As a result, creating LLMs in a multimodal setting presents both a technological challenge and an innovative chance to rethink the way that rare illness research and treatment are perceived.

Cross-disciplinary cooperation between AI researchers, physicians, geneticists, and bioinformaticians will be necessary to overcome these obstacles. According to Hasani et al. , this kind of cooperation is essential to guaranteeing that developments in AI technology are clinically significant and satisfy patient requirements.

Critical issues in data standardization, ethical issues, and converting AI-driven solutions into practical applications can be resolved by active cross-disciplinary collaboration. In order to emphasize therapeutic outcomes, Denecke et al. \cite{denecke2024} advocate collaborative approaches, which would guarantee innovation but, more crucially, the usefulness and accessibility of AI technologies. Ou et al. \cite{ou2024} assert that interdisciplinary approaches have the best chance of breaking through bottlenecks including, among other things, clinical interpretation of outcomes and standardized dataset production in numerous modalities. 

By encouraging cooperation and combining a variety of skills, the research community may expand the realm of what is feasible using LLMs in rare disease scenarios.
\newpage

\bibliography{biblio}

\end{document}